\documentclass{article}
\usepackage{ijcai16}
\usepackage{times}
\usepackage{joe-customizations}

\usepackage{placeins}
\usepackage{float}
\usepackage{dblfloatfix}

\usepackage{CJKutf8}
\newcommand{\kanji}[1]{\begin{CJK}{UTF8}{min}#1\end{CJK}}

\title{Teaching natural language to computers}

\author{~\\[-1\baselineskip]
\begin{tabular}{cc}
\textbf{Joseph Corneli} & \textbf{Miriam Corneli} \\
Department of Computing & Former Senior English Language Fellow\\
Goldsmiths College, University of London & (Nepal / Sri Lanka), US Department of State\\
j.corneli@gold.ac.uk & languagecoach.miriam@gmail.com\\
\end{tabular}
}

\begin{document}
\maketitle

%% Here, in fact, is the abstract.  Feel free to try
%% making a few "practice" changes to the text of the abstract, and see what
%% appears in the box at right.  Then you can get going on whatever changes
%% you like.  OK, no more comments for now.

\begin{abstract}
``Natural Language,'' whether spoken and attended to by humans, or
  processed and generated by computers, requires networked structures
  that reflect creative processes in semantic,
  syntactic, phonetic, linguistic, social, emotional, and cultural modules.
  Being able to produce novel and useful behavior following repeated
  practice gets to the root of both artificial intelligence and human
  language.  This paper investigates the modalities involved in
  language-like applications that computers -- and programmers -- 
  engage with, and aims to fine tune the questions we ask to
  better account for context, self-awareness, and embodiment.
\end{abstract}

\section{Introduction}

In the genesis of intelligent computational systems, one often
observes programs that write before they can read, compose before they
can listen, and paint before they can see.
However, the most successful systems in poetry, music, and visual art
are indeed perceptually aware -- and derive significant benefits from
that ability
\cite{kurzweil1990age,screenelearning,colton2015painting}.

We will argue that the bar by which we judge computational creativity
in text generation -- and appreciation -- can and must be raised, in
order to build systems that we can meaningfully communicate with.  Our
paper takes the form of a necessarily high-level sketch, supported by
hand-crafted examples that draw on both standard and custom software.
Most of these examples concern computer poetry, but our aim is not to
present a technical or aesthetic achievement.  Rather, we use the
examples to survey the limitations of current systems and to indicate
some potentially novel approaches.

Specifically, we highlight three interrelated thematic areas that we think
will repay effort.
\paragraph{(Limited) Contextual Understanding}  Here, we are concerned with
  what makes a response to some circumstance meaningful.  This certainly
  requires \emph{context} \cite{ogden1923meaning}.  For example, a response in a conversational dialog
  generally considers the previous elements of the exchange, and perhaps also previous exchanges,
  and elements of a shared culture.  This understanding does not need to be, nor in general
  can it be, ``complete.''   Furthermore, in many cases, the reader or listener will hear meanings that were not there originally  \cite{veale2015game} -- however, the interpolation of meaning by the reader cannot always be relied upon.
  Later in this paper we include an example of a
  computer-generated poem that is essentially just verbose babbling
  fitted to a predefined template.  This poem was previously reviewed
  by a published poet, and it does not stand up well to critical
  scrutiny \cite{corneli2015computational}.  The poem misses any sense
  of ``why'' -- and the program that generated this poem would not be
  able to offer what the Proven\c{c}al poets called a \emph{razo}
  \cite[p.~51]{agamben1995idea}, that is, an exposition of the poem.
\paragraph{(Limited) Self-awareness} Here, we focus exclusively on the ability to reason about creativity as
  a process.  Computer programs often have limited metadata
  about their software processes, for example the ``signatures'' of
  functions, specifying the types of input data that the functions will accept,
  or ``contracts'' specifying preconditions, postconditions, and invariants.
  Along with these representations often comes some limited ability to reason formally about code.
  But few if any contemporary software systems would be able to convincingly make
  sense of a simple writing prompt, or adapt a dialogical process of response in order to reach an agreement, or
  respond to feedback from a critic in order to produce a better poem.
  In the future, such computational abilities with language may be commonplace.
  These abilities may depend on fairly profound epistemic features, for example,
  the computer might need to recognize when its ``knowledge'' is uncertain, and proceed
  accordingly -- perhaps asking for help, or making multiple
  generative attempts in parallel and assessing which one
  works better relative to its contextual understanding.
\paragraph{(Limited) Embodiment} We will consider program flowcharts
\cite{charnley2014flowr} as the primary framework with which to
describe the computer's ``process'' layer.  Whether embodied
as a flowchart or a Von Neumann machine or something else,
computational processes are also physical processes.  The manipulations of the nodes and arrows
of a flowchart or of some other collection of physical objects, like the flowchart's corresponding script,
or the words of a poem, can (potentially) be thought about with respect to its gestural content.  This definition of
\emph{gesture} due to the 12th Century theologist Hugues de
Saint-Victor, quoted in \cite{mazolla2016melting}, shows the
connection with our other themes:
\emph{Gestus est motus et figuratio membrorum corporis, ad omnem agendi et habendi modum.} 
Mazolla glosses this as follows:
\begin{quote}
Gesture is the movement and figuration of the body's limbs with an
aim, but also according to the measure and modality proper to the
achievement of all action and attitude.
\end{quote}
Introducing this quote requires us to make one significant caveat.  Whereas humans tend to perceive ourselves as relatively free beings, able to act according to a purpose and even to choose which purpose to serve, we regard computers in a very different light.  At best, a computer can be programmed to optimize its behavior relative to some constraint.  This perspective does not sit well with the typical understanding of the English word ``aim.''\footnote{It would appear that Mazolla freely introduced this concept, instead of sticking with the more literal ``agenda.''}  The point to make here is that a \emph{computational system} is understood relative to an operating environment, and its behavior is worked out relative to that environment.  Under some circumstances we would call this process ``programming,'' and under somewhat different circumstances we would describe it as ``self-programming'' or ``automatic programming.''  In short, it is not necessary to attribute intention to the computer, but -- once again -- it is necessary to think about its behavior in context.

% Gestus est motus et figuratio membrorum corporis, ad omnem agendi et habendi modum.
% Gesture is the movement and figuration of the body's members,
% And the configuration of the gesture is the movement of the members of the body, to carry out and to have the way of all.  

The remainder of the paper is structured as follows:
First, we will explore these themes from a computing standpoint, developing
a technical sketch rather than a formal system description.
Then we turn to a discussion that evaluates this technical
sketching from the point of view of the second author, an English as a Foreign Language teacher
with a prior background in consciousness studies.  Finally, we draw on
this discussion to outline a plan for future computational
experiments centered on making sense of language.

\section{Motivation} \label{sec:motivation}

Oscar Schwartz offers the following framing:

\begin{quote}
Can a computer write poetry? This is a provocative question. You think
about it for a minute, and you suddenly have a bunch of other
questions like: What is a computer? What is poetry? What is
creativity? \cite{can-computer-write-poetry}
\end{quote}

Computer poetry may sound like a bit of a lark -- after all, it's not
clear that anyone really needs it.  Nevertheless, asking these
questions about poetry begins to suggest a way of working with language that has wider implications.

Consider Turing's idea that machines ``would be able to converse with
each other to sharpen their wits'' \cite{turing-intelligent}.  This
could be realized as a \emph{Q\&A site specifically for computers}.
The discussions could address all manner of practical concerns,
for example, those arising for bots that are engaged in code or
editorial review tasks.  A reputation system and web of trust could be used
to maintain quality control.
If the participating computational systems had sufficient abilities
with natural language, this system could be bridged into a Q\&A site
that is in everyday use by human beings.
However, before a computational system would be useful in any Q\&A
context, it would presumably need to be able to be able to model
the meanings of the questions that are being asked reasonably well,
and also to be able to compose meaningful responses.  For now, we will side-step the
Chinese Room-style question of whether the system ``really'' understands
what it is processing \cite{bishop2004view} and focus on the more
applied question: how would meaningful responses occur?
A high-level outline could be something like this \cite{corneli2015computational}:
(1) Read and understand the ``prompt'' to a sufficient degree; (2) Compose
a response that ``makes sense''; (3) Criticize the response along
various dimensions, for instance, does it read well, does it tell a
story or develop a character?; (4) Consider how it might be improved.
This outline is based on an established process that groups of
creative writers use to critique and revise poetry or literary works
for publication.  

When we turn to computer generated text -- say, poetry, to be concrete
-- in addition to examining the generated poem, a sophisticated
audience can also examine the \emph{process} whereby the poem was
generated, and read the product against this process (or vice versa).
Indeed, the computer can create both poem and process, the later via
automatic programming -- and offer its own assessments of both as
well, as long as it can make sense of the success criteria.

\subsection{Related work}

Natural language processing often begins with a grammar.  If none is
available, it may be induced, for instance by using compression
techniques \cite{wolff1988learning}.  Both in older
\cite{redington1997probabilistic} and quite recent work
\cite{hermann2015teaching}, statistical and neural network approaches
to corpus-based language understanding have shown strong potential for
developing ``reasoned'' ways of thinking about linguistic structure
without the usual grammatical assumptions.  Corpus methods help to
understand the patterns in the way people use language, and the
creative potential of unexpected word combinations
\cite{hoey2005lexical}.

One example cited by \citeauthor{hoey2005lexical} is Tennyson's famous
line \emph{Theirs is not to reason why}.  Here, the word \emph{reason}
is used with its verb sense, rather than with the noun sense that most
readers would expect based on their prior experience with the two-word
phrase \emph{reason why}.  This unique feature makes the line
memorable and interesting.  The psycholinguistic properties of the
broader phenomenon of ``lexical priming'' have been extensively
studied \cite{pace2013concept}.  One empirical result is that priming works differently for native speakers and for  non-native ESL speakers, insofar as native speakers are more affected by binding of words within formulas, whereas someone learning a new language tends to only recognize the strings that they have encountered before.

The ascendant status of data-driven methods in
natural language processing does not obviate symbolic AI, which
continues to be useful for work with specialist languages.  For
computer programming languages in particular, techniques for
\emph{reflection} and, in the case of LISP, \emph{homoiconicity}
(i.e., treating code as data) make it possible to write computer
programs that reason about, write, or adapt computer programs.

Artificial languages have been used in video games in creative ways, but not, as yet, for functional communication with or between
non-player characters.  Multi-agent systems have, however, been used
in poetry generation.  One example system creates poems based on
repeated words and sounds driven by a model of agents' emotional
states \cite{kirke13}, inspired by earlier work in music
\cite{kirke2011application}.  As for computer programs that read
poetry, this is typically limited to reading (and mimicking)
surface style, without extending to \emph{meaning}
\cite{carslisle2000comments} -- even if some readers were fooled.
Without knowing what's in a poem, it seems difficult to be other than
superficial.

Just how far the ``surface'' goes is a question much discussed by
poets and translators of poetry.  Red Pine, translator of the 13th-14th Century Chinese poet and Buddhist monk Stonehouse
(\kanji{石屋}) wrote as follows \cite[p.~xxiv]{pine2014mountain}:
\begin{quote}
I don't know how others do it, but when I've tried to think of a metaphor for what I go through, I keep coming up with the image of a dance.  % I see the poet dancing, but dancing to music I can't hear.  Still, I'm sufficiently enthralled by the beauty of the dance that I want to join the poet.  % And so I try.  And as I do, I try not to step on my partner's feet (the so-called literal or accurate translation) but I also try not to dance across the room (the  impressionistic translation or version -- usually by someone who doesn't know the poet's language). 
[\,\ldots]
I try to get close enough to feel the poet's rhythm, not only the rhythm of the words but also the rhythm of the poet's heart.
\end{quote}
A typical approach to poetry generation might take Stonehouse's corpus and notice that he often writes about clouds and mountains and plants, and attempt to generate a poem ``in the style of Stonehouse,'' referencing some of these typical concepts and aiming to get the number of syllables and the grammatical structure right.  However, there is little doubt that a reader with an ear for Chinese poetry or some familiarity with the ideas of Chan Buddhism would recognize these \emph{ersatz} attempts for what they are: ``dead words'' \cite[p.~xxiii]{pine2014mountain}.  More interesting computer poems in a somewhat related genre (Japanese \emph{haiku}) have been created by programmers working from the premise  \cite[p.~2497]{rzepka2015haiku} that a reader's interest in a haiku stems from:
\begin{quote}
feeling that the poet understands a situation and that we can mentally agree with what she/he (or maybe \emph{it}) shares with us.
\end{quote}

The  challenges posed by computer poetry serve as a
point of departure. 
``Poetry exercises are used to allow learners to explore the
complexities of English'' \cite{graph-theoretic-poetry} -- or another language -- and the contexts in which it is meaningful.

\section{Exploration}

Here are 10
% \href{http://stackoverflow.com/questions/5917576/sort-a-text-file-by-line-length-including-spaces}{short}
short examples of writing prompts excerpted from the book ``642 Things to Write About'' \cite{things-to-write-about}:
\begin{enumerate}[itemsep=.1cm,leftmargin=1cm]
\item What can happen in a second.
\item The worst Thanksgiving dish you ever had.
\item A houseplant is dying. Tell it why it needs to live.
\item Tell a story that begins with a ransom note.
\item Write a recipe for disaster.
\item If you had one week to live\ldots.
\item What your desk thinks about at night.
\item The one thing you are most ashamed of\ldots
\item Describe your best friend.
\item Describe Heaven.
\end{enumerate}

In order to create a computer-generated response to any of these prompts,
in addition to understanding what the prompt is saying, some further understanding of the topic
is required.  The response itself will have various standard
features.  Many of the following features of \emph{stories} are found (with
minor adaptations) in poetry, and other kinds of writing:

\begin{quote}
A story is not a modular presentation of ideas but a multi-layered
work consisting of interdependent characters, plot elements, and
settings. \cite{kim2014ensemble}
\end{quote}

Let us consider, then, a simple theory of stories and storytelling,
using the prompts above as our domain.  One suitable theory would involve a micro-world containing:
\emph{A Scenario}, \emph{A Narrator}, \emph{An Audience},
\emph{A Beginning}, \emph{A Middle}, and \emph{An End}.
Consider that -- with respect to the ``Thanksgiving dish'' prompt --
the computer has presumably never tasted food of any kind.  It could,
however, ``imagine'' a scenario in which there is a character who eats
a Thanksgiving dish.

\subsection{An example \emph{Scenario}}
\label{sec:orgheadline3}

The scenario could be represented with various relations:

\begin{quote}
Squanto \texttt{memberOf} Patuxet tribe, Patuxet tribe
\texttt{hasCardinality} 1, Thanksgiving \texttt{isa} event,
Thanksgiving \texttt{hasHost} Pilgrims, Thanksgiving \texttt{hasGuest}
Squanto, Thanksgiving \texttt{hasFood} eel, eel \texttt{hasCondition}
burnt
\end{quote}

Naturally, this might be extended with further information as that
comes to light; and in practice we might use a more robust formalism.

\subsection{The other components}
\label{sec:orgheadline4}

The \emph{Narrator} would walk through the scenario and say what's
there.  On a metaphorical level, the narrator's role is somewhat
similar to the way a virtual camera moves through a 3D simulation in
order to create a film.  However, the \emph{Narrator} needs to
consider the \emph{Audience} in order to be effective.  As indicated
in the quote from \citeauthor{kim2014ensemble}, above, the data in the
\emph{Scenario} needs to be structured when telling the story.

Here is one possible presentation of the scenario, embellished with
some facts, fictions, and local color, and combined into several
sentences that flow reasonably well.

\begin{quote}
Squanto was the last surviving member of the Patuxet tribe.  He
attended the first Thanksgiving with the Pilgrims wearing a new
buckskin jacket.  One of the foods that was served was eel, but it had
been rather badly burnt and Squanto didn't find it to his liking.
\end{quote}

This text manages to include a range of emotionally evocative, thought
provoking, character establishing, and sensory language (``last
surviving member,'' ``first Thanksgiving,'' ``new buckskin jacket,''
``rather badly burnt,'' ``didn't find it to his liking'') -- and at
least one unexpected word combination (``Thanksgiving\ldots\ eel'').
It also has a discernible \emph{Beginning}, \emph{Middle}, and
\emph{End}.  It seems appropriate to call it a \emph{story}.

\subsection{How to come up with stories?}
\label{sec:orgheadline6}

Let's start with a parse:

\begin{lstlisting}[breaklines]
(TOP
 (NP 
  (NP (DT The) (JJS worst)
      (NNP Thanksgiving) (NN dish))
  (SBAR (S (NP (PRP you)) (ADVP (RB ever))
           (VP (VBD had)))) (. .)))
\end{lstlisting}

\noindent Here are some associated word meanings from WordNet:\footnotemark

\begin{center}
\begin{tabular}{lp{.25\textwidth}}
\textbf{Word} & \textbf{Gloss from WordNet}\\
\hline
\texttt{(DT The)} & \texttt{determiner}\\
\texttt{(JJS worst)} & (superlative of `bad') most wanting in quality or value or condition\\
\begin{tabular}[t]{@{}l@{}}\texttt{(NNP}\\\texttt{~Thanksgiving)}\end{tabular}& commemorates a feast held in 1621 by the Pilgrims and the Wampanoag\\
\texttt{(NN dish)} & a particular item of prepared food\\
\texttt{(PRP you)} & \texttt{pronoun}\\
\texttt{(RB ever)} & at any time\\
\texttt{(VBD had)} & serve oneself to, or consume regularly\\
\end{tabular}
\end{center}

There are some other interesting word senses available, and knowing
which one to pick, or how (and how much) to combine various senses
seems like a bit of an art form.  Should we consider \emph{a short
  prayer of thanks before a meal} when thinking of ``Thanksgiving''?
For now, we will say a short prayer and tentatively assume that a
\verb|(NN dish)| is what you eat, rather than what you eat off of.
``Lexical priming'' \cite{hoey2005lexical} techniques would 
help make the relevant distinction here.
%%  (Had the prompt referred to a ``commemorative dish'' we
%% would have been better off making the opposite assumption.)

But supposing we get this far, now what?  We've moved from one
sentence to several quasi-sentences, without getting that much closer
to a ``\emph{Scenario}'' like the one envisaged above.  One
possibility is that the WordNet expansion could be sufficient give us
relevant keywords and phrases, from which a small corpus could be
built (e.g., by doing a web search for the glosses) and then mined to
learn relations between the items in that corpus.  Alternatively or
additionally, these meanings might be connected to a pre-computed
model of linguistic meaning drawing from a much larger background corpus
\cite{mcgregorwords}.

%% .  Then,
%% graph-based NLP methods could be used to ``learn'' the relations
%% between items in that corpus.
\href{http://www.cs.toronto.edu/~tsap/publications/icdm02.pdf}{Mining
  significant associations from large scale text corpora} is something
people have explored in various ways.  Finding subject-verb-object triples, in
particular,
\href{https://www.academia.edu/5349637/TRIPLET_EXTRACTION_FROM_SENTENCES}{is
  a popular method}: one well-known algorithm is presented by
\citename{rusu2007triplet}.  This is sometimes called ``building a
semantic model.''  Other more sophisticated approaches might draw on 
associations with a pre-existing ontology \cite{kiryakov2004semantic} -- the
particular benefit of the \citeauthor{rusu2007triplet}~approach is that it can be implemented
using a simple parsing-based method.  The basic theme of building semantic models
of text goes back to
\citename{quillian1969teachable} -- about whom there will be more to
say later.
For now, we just remark that in addition to expanding the writing
prompt, we may also want to draw on some ``stock'' associations stored
in a background knowledge base.  We 
illustrate the method from \citeauthor{rusu2007triplet}~by applying it to
the beginning of the novel \emph{Frankenstein} (Table \ref{frankenstein-triples}).

\begin{quote}
You will rejoice to hear that no \textsl{disaster} has accompanied the
commencement of an enterprise which you have \textsl{regarded} with such evil
\textsl{forebodings}.\hspace{.7mm}\ldots 
\cite{frankenstein} (emphasis added)
\end{quote}

\begin{table}[H]
{\centering
\fontshape{sl}\selectfont
\begin{tabular}{llll}
\normalfont{1}. & disaster & regarded & forebodings\\
\normalfont{2}. & yesterday & increasing & confidence\\
\normalfont{3}. & London & fills & delight\\
\normalfont{4}. & feeling & understand & feeling\\
\normalfont{5}. & breeze & gives & foretaste\\
\end{tabular}

\par}
\vspace{-.1cm}
\caption{5 triples extracted from \emph{Frankenstein}\label{frankenstein-triples}}
\end{table}

\footnotetext{NB., WordNet contains no entries for determiners or pronouns.}

\FloatBarrier

As far as text understanding goes, the result in Table
\ref{frankenstein-triples} is not particularly encouraging.  However,
even a low-fidelity database of background knowledge would allow us to extend the story.
Perhaps Squanto would decide that the burnt eel \textsl{gives} a
\textsl{foretaste} of things to come.  To make this association, we
might use methods similar to the ones used to reason about
ConceptNet triples -- which have been employed to good effect in text and concept
generation within the FloWr framework \cite{Llano2016}, which is described below.

\subsection{Can FloWr flowcharts be used to solve the composition problem?}
\label{sec:orgheadline8}

\begin{figure*}[t]
\vspace{.2cm}
{\centering
\includegraphics[width=.8\textwidth]{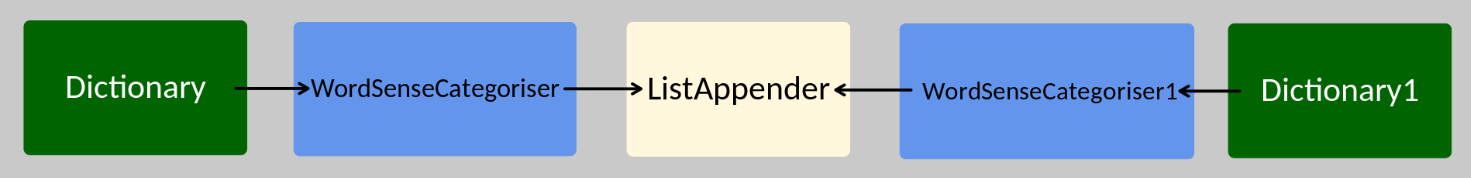}\\[.1cm]
\includegraphics[width=.8\textwidth]{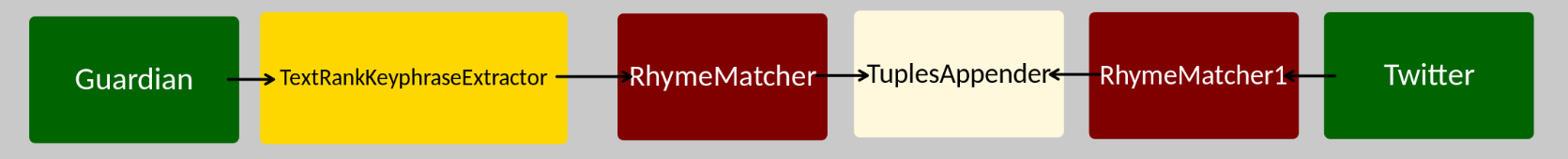}

\par}

\vspace{-.3cm}
\caption{Two simple FloWr flowcharts \label{simple-flowcharts}}
\end{figure*}

FloWr \cite{charnley2014flowr} is a flowcharting system with basic
text processing abilities.  Metadata describing the ``ProcessNodes''
from which FloWr's flowcharts are formed can be used to pose and solve
simple automatic programming problems.
A writing prompt like ``worst Thanksgiving dish'' can be interpreted as
a constraint -- or, more broadly, a fitness function -- that steers
the generative process, and trickles through to guide the choice of
functional components and, eventually, words.  By ``fitness
function,'' we understand that text may be composed iteratively, and
improved along the way, relative to some context of evaluation.  The
simple examples in Figure \ref{simple-flowcharts} illustrate problems
that can be solved quite straightforwardly:
\begin{quote}
``Give me a list of mixed \emph{adjectives} and \emph{nouns}.''
\end{quote}
(the italicized terms are ``independent'' variables that tell the
\texttt{WordSenseCategoriser} node how to behave); or
\begin{quote}
``Give me a list of \emph{5} rhyming couplets built of text from The Guardian and Twitter mentioning `\emph{eels}'.''
\end{quote}

It may make sense to include considerably more abstraction in the
description of larger and more complex flowcharts.  For example, a
flowchart discussed by \citename{corneli2015computational} includes 28
nodes and generates the following poem (and others that are similar):

{
\fontshape{it}\selectfont
\begin{verse}\label{demon-dog}
\hspace{-.5em}Oh dog the mysterious demon\\
\hspace{-.5em}Why do you feel startle of attention?\\
\hspace{-.5em}Oh demon the lonely encounter\\
\hspace{-.5em}ghostly elusive ruler\\
\hspace{-.5em}Oh encounter the horrible glimpse\\
\hspace{-.5em}helpless introspective consciousness\\
\end{verse}
}

\begin{table*}[t]
{\centering
\begin{tabular}{lll}
number \texttt{N}                      & \texttt{Dictionary} & list of words of length \texttt{N}\\
list of words, word sense             & \texttt{WordSenseCategoriser} & list extracting words with the word sense\\
input string(s)                      & \texttt{TextRankKeyphraseExtractor} & list of key phrases extracted from the strings\\
phrases, number \texttt{M} of phonemes & \texttt{RhymeMatcher} & tuple of couplets with \texttt{M} rhyming phonemes\\
number \texttt{N}, word                & \texttt{Twitter} & some \texttt{N}  tweets containing the word\\
tuples                                 & \texttt{TuplesAppender} & a list combining the tuples\\
lists                                  & \texttt{ListAppender} & a list combining the lists\\
\end{tabular}

\par}
\vspace{-.2cm}
\caption{Triples describe the functional mapping from input to output for selected FloWr nodes \label{tab:flowr-functional-relationships}}
\end{table*}

Would the most succinct description of the flowchart be an
approximately 28 clause sentence that is \emph{equivalent} to the
flowchart?  Perhaps, in the current case, everything can be
compressed down to the following template (fixed
for this flowchart), and potentially further with optimizations:

{\footnotesize
\fontshape{it}\selectfont
\begin{verse}\label{demon-dog-behind-the-scenes}
\hspace{-.5em}Oh THEME the COLLOCATE SIMILAR\\
\hspace{-.5em}Why do you feel INVERSION of DESIRED?\\ 
\hspace{-.5em}Oh \hspace{-.4em} SIMILAR \hspace{-.4em} the \hspace{-.4em} COLLOCATE-OF-SIMILAR \hspace{-.4em} SIMILAR-TO-SIMILAR\\
\hspace{-.5em}SIMILAR-TO-COLLOCATE \hspace{-.3em} SIMILAR-TO-COLLOCATE\hspace{.02em}${^\prime}$\\
\hspace{1em} COLLOCATE-OF-SIMILAR\\
\hspace{-.5em}Oh\hspace{-.2em} SIMILAR-TO-SIMILAR\hspace{.02em}${^\prime}$\hspace{-.2em} the\hspace{-.2em} \hspace{-.5em}COLLOCATE-OF-SIMILAR-TO-SIMILAR SIMILAR-TO-SIMILAR-TO-SIMILAR\\
\hspace{-.5em}SIMILAR-TO-COLLOCATE SIMILAR-TO-COLLOCATE\\
\hspace{1em} COLLOCATE-OF-COLLOCATE\\
\end{verse}
}

The connection between this template, its instantiation, and the
putative prompt, ``Write a poem about an old dog who is afraid of
attention,'' is tenuous at best.  Nevertheless, reasonable hope
exists for future work that would generate models -- and a
\emph{Narrator} -- tailor-made to a given prompt.

FloWr's ProcessNodes define a micro-language denoting the available
ways in which the system can transform input data to output data
(Table \ref{tab:flowr-functional-relationships}).  In a meaningful
expansion of a given prompt, many choices would have to be made.
New flowcharts built in
response to writing prompts or other contextual data would constitute
the system's core ``learnings.''  In short, poetry and process need to
be thought about together.

\section{Discussion}

In this section, we will briefly review human language learning from a
second language teaching perspective, and then draw comparisons with
the foregoing description of a hypothetical computer language learner.

First, why are primary language learning and second\slash foreign
language learning often considered separately?  One of the biggest
differences between the two cases is that after learning a first
language, ``neural pathways'' have been set down, so that second
language information has to be encoded ``along with'' or ``beside''
the first language pathways.  New neural connections have to be formed
to maintain the memory of the second language, whereas the first
language has been implanted quite thoroughly.  So when learning verbs,
nouns, etc., the person (child, adult) learns or has learned, for
instance, ``\emph{stand up!}''~first and then that
``\emph{levantate!}''~means a similar thing, in Spanish, or that
``\emph{chair}'' = ``\emph{silla}.'' As they progress in the second
language, what we find is a process that linguists call
``interlanguage'' \cite{selinker1972interlanguage}, or a language that
is not quite English, and not totally Spanish yet, because of the
variations in the two languages in structure, phonemes, morphemes,
allophones, semantics, and so on.

Language is also very multi-modal, and if we think again in terms of
neurons and brain pathways, the word ``dog'' not only brings up images
of a furry canine animal but all kinds of other associations.  Pet
dogs.  Large and small dogs.  Dogs with different color fur (visual
cortex).  \emph{Dog} as a verb (motor cortex associations).  The smell
of wet dogs (olfactory bulb).  Fear of dogs (limbic system).  Favorite
pet dogs you have loved in your life (emotions).  The spelling or
sound of the word D-O-G as opposed to D-A-W-G (auditory and visual
cortex).  If you are from certain countries perhaps even dog meat.
Imagining a ``demon dog'' will invoke networks running all over the
brain \cite{Schlegel01102013}.  Now what about computers, do they have
a similar symbolic or multi-modal operating ability?  People, as they
become literate, learn to write and sound out letters at the same time
-- and the foundations of first language learning draw heavily on a
sensory-motor channel \cite{iverson1999hand,hernandez2013bilingual}.

Computers may be said to have a first language with several dialects,
highly constrained by grammar -- namely, bytecode and programming
languages.  The flowcharts described above begin to recover a degree
of multi-modality, and a process orientation that is similar in
certain partial respects to sensory-motor experience.  We might also think of flowcharts as akin to neurons and
cortexes, as above.  After all, there is a kind of embodiment even in
the brain -- it is an active and evolving organ \cite{doidge2007brain}.
Flowcharts are rather different from classical neural network models, but one common feature
is that they would need to be ``trained'' if they are to understand
and express language.

Again, poetry could be part of the way forward.  Exercises from a
book like \emph{Writing Simple Poems: Pattern Poetry for Language
  Acquisition} \cite{holmes2001writing} might be used to teach
computers as well as humans.  Note that grammar and poetry are very
different, and perhaps complementary.  Thus, for example, a ``Learn
English!''~notebook found in Japan with \emph{Subject -- Verb --
  Direct Object -- Indirect Object -- Prepositional Phrase --
  Adjectival Phrase} written at the top of each page offers a useful
rubric for Japanese students, since the verb comes at the end of the
sentence in Japanese (SOV).  But grammar will only get you so far.
Consider Chomsky's famous nonsense statement, ``Colorless green ideas
sleep furiously.''  A poetic gesture -- like prefixing that phrase
with a description of ``planted tulip bulbs'' -- is able to make some
sense out of nonsense by adding
context!\footnote{\url{http://www.linguistlist.org/issues/2/2-457.html\#2}}

Social context is also likely to be relevant: peer learning is
very useful for human language learners \cite{ranciere1991ignorant,raw2014paragogy}.
One idea would be to adapt the Q\&A model mentioned in Section \ref{sec:motivation} as a ``social'' site about poetry.

\section{Future work}

Quillian's ``The Teachable Language Comprehender: A Simulation Program
and Theory of Language'' \shortcite{quillian1969teachable} took the
novel -- and fundamental -- approach of understanding things in such a
way that new understandings could be added directly to its knowledge
base.\footnote{In practice, ``While the monitor can add TLC's encoded
  output to the program's memory, the program itself makes no attempt
  to do so, nor to solve the problems inherent in doing so''
  \cite[p.~473]{quillian1969teachable}.}  When reading a piece of
text, the TLC program would search its memory for related information
that it could use to make sense of the input.  More specifically, a
given text would be expanded using ``form tests'' which extracted
meaningful pieces of the text, and connected these to items stored in
memory.  Quillian writes that ``ultimately, a human-like memory should
relate descriptive knowledge of the world to perceptual-motor
activity, in a manner like that indicated by Piaget'' -- but deems
this to be ``far beyond our present scope''
\cite[p.~474]{quillian1969teachable}.

Future research might use FloWr to develop a TLC-like library of
``form tests'' and generative tools that would add a multi-modal
aspect to knowledge representation.  
%% Especially in the domain of poetry, learning a text is not just a
%% matter of understanding the content, but of getting ``a feel'' for
%% the way the poem works.
To be sure, a flowchart-based representation of poetic process would
be quite different from the embodied sensory-motor experience of
humans.  Nevertheless, computational processes that allow us to model
text generation contextually, procedurally, and gesturally can help to
understand the way linguistic meaning comes to be.  This is not
something we can readily learn from parsing, corpus-based modeling, or
grammar-based text generation alone.
There is exciting potential for future experiments with natural
language that strives to capture and express shades of meaning and the
``feel'' of the language.  Experimentation is necessary: if we have
learned anything about language, it is that ``learners should be
motivated to speak bravely'' \cite{wang2014developing}.

\section{Acknowledgments}
Joseph Corneli's work on this paper was supported by the Future and
Emerging Technologies (FET) programme within the Seventh Framework
Programme for Research of the European Commission, under FET-Open
Grant number 611553 (COINVENT).
Thanks to Teresa Llano for conversations about her work on the ``demon
dog'' poem, quoted above.

\bibliographystyle{named}
\bibliography{iccc}

\clearpage

\end{document}